\documentclass[11pt]{article}

\usepackage[final]{acl}

\usepackage{times}
\usepackage{latexsym}
\usepackage[T1]{fontenc}
\usepackage[utf8]{inputenc}
\usepackage{microtype}
\usepackage{graphicx}

\usepackage{amsmath}
\usepackage{amssymb}
\usepackage{booktabs}
\usepackage{array}
\usepackage{float}
\usepackage{xcolor}
\usepackage{colortbl}
\usepackage[most]{tcolorbox}
\usepackage{url}
\usepackage{enumitem}
\usepackage{arydshln}

\usepackage{listings}
\lstdefinestyle{prompt}{
  basicstyle=\ttfamily\scriptsize,
  breaklines=true,
  breakindent=0pt,
  columns=fullflexible,
  keepspaces=true,
  xleftmargin=0pt,
  xrightmargin=0pt,
}

\definecolor{visioncolor}{RGB}{194, 97, 12}      
\definecolor{textcolor}{RGB}{0, 121, 107}         
\definecolor{bothcolor}{RGB}{103, 58, 183}        
\definecolor{visionbg}{RGB}{255, 243, 224}
\definecolor{textbg}{RGB}{224, 247, 244}
\definecolor{bothbg}{RGB}{243, 229, 255}

\definecolor{findingbg}{RGB}{237, 244, 252}
\definecolor{findingborder}{RGB}{66, 133, 244}
\definecolor{bestgreen}{RGB}{226, 240, 226}
\definecolor{takeawaybg}{RGB}{232, 245, 233}
\definecolor{takeawayborder}{RGB}{56, 142, 60}
\definecolor{headergray}{RGB}{240, 242, 245}
\definecolor{stripegray}{RGB}{248, 249, 251}
\definecolor{highdis}{RGB}{253, 226, 220}   
\definecolor{lowdis}{RGB}{220, 242, 220}    

\definecolor{promptbg}{RGB}{249, 249, 249}
\definecolor{promptborder}{RGB}{180, 180, 180}
\definecolor{prompttitle}{RGB}{60, 60, 60}

\newtcolorbox{finding}{%
  colback=findingbg,
  colframe=findingborder,
  boxrule=0.5pt,
  arc=2pt,
  left=6pt, right=6pt, top=4pt, bottom=4pt,
  fontupper=\small
}

\newtcolorbox{takeaway}{%
  colback=takeawaybg,
  colframe=takeawayborder,
  boxrule=0.5pt,
  arc=2pt,
  left=6pt, right=6pt, top=4pt, bottom=4pt,
  fontupper=\small
}

\newtcolorbox{contribbox}{%
  colback=white,
  colframe=findingborder,
  boxrule=0pt,
  borderline west={3pt}{0pt}{findingborder},
  arc=0pt,
  left=8pt, right=6pt, top=4pt, bottom=4pt
}

\newtcolorbox{promptbox}[1][]{%
  enhanced,
  breakable,
  colback=promptbg,
  colframe=promptborder,
  boxrule=0.4pt,
  arc=2pt,
  left=6pt, right=6pt, top=4pt, bottom=4pt,
  fonttitle=\small\sffamily\bfseries,
  coltitle=prompttitle,
  colbacktitle=white,
  toptitle=3pt, bottomtitle=3pt,
  title={#1},
  before skip=6pt,
  after skip=6pt
}

\title{Text, Pixels, or Both? Evaluating Input Representations for Multimodal Document QA}

\author{
Nikhil Reddy Pottanigari\thanks{\ \ Equal contribution.} \hspace{1.5em} Sepideh Kharaghani\footnotemark[1] \hspace{1.5em} Saverio Vadacchino \\[0.15em]
\bfseries Alejandro Posada \hspace{1.5em} Kurt MacDonald \hspace{1.5em} Ying Zhang \\[0.4em]
\bfseries ServiceNow Canada \\
\texttt{\{nikhilreddy.pottanigari, sepideh.kharaghani,} \\
\texttt{saverio.vadacchino, alejandro.posada,} \\
\texttt{kurt.macdonald, yin.zhang\}@servicenow.com}
}

\newcommand{\vision}{{\color{visioncolor}\textsc{Image}}}
\newcommand{\textrep}{{\color{textcolor}\textsc{Text}}}
\newcommand{\both}{{\color{bothcolor}\textsc{Text{+}Image}}}
\newcommand{\oracle}{\textsc{Oracle}}
\newcommand{\allpages}{\textsc{Full}}
\newcommand{\best}[1]{\cellcolor{bestgreen}\textbf{#1}}
\newcommand{\acc}{\mathrm{Acc}}

\newcommand{\dis}{\mathrm{Dis}}

\begin{document}

\maketitle

\begin{abstract}
Every document QA system begins with a choice that is rarely studied on its own: whether to feed
the model page images, extracted text, or both. We isolate this choice, holding the prompt, judge,
and scoring pipeline fixed, across four commercial model endpoints, two corpora, and two context
regimes (gold evidence pages and the full document). On documents that fit the image budget,
page images lead on accuracy at every document length on both corpora, but this advantage
carries a growing latency and cost premium: text latency stays roughly flat as documents lengthen
while image latency rises steadily. Text and images also fail on different questions, with exactly
one representation correct on \textbf{19--25\%} of items across the reported cells, so neither
subsumes the other. Exploiting this complementarity, a lightweight TF-IDF router that reads only
the question text gains \textbf{2.6 points} over always-text while cutting median latency
\textbf{30\%} relative to always-vision, on a document-disjoint held-out split. 
\end{abstract}
\section{Introduction}

Before a document QA system can answer a question, a simple preprocessing choice
sets how the document reaches the model: as page images, as extracted text, or both
\citep{zhang2024document}. A knowledge-base article may encode evidence in a screenshot, diagram, or spatial arrangement that OCR text preserves only partially. Sending every page as an image retains this visual evidence but increases input size and inference latency on long documents. Combining both modalities can be worse than either
channel alone, depending on the model.

This choice affects accuracy and latency together. Always sending images for a long report, or text for a screenshot-heavy knowledge base, can hurt answer quality or increase inference time. However, representation is often evaluated together with changes to the model, prompt, retrieval method, or evaluator, making its contribution difficult to identify.

\begin{figure}[t]
\centering
\includegraphics[width=0.97\columnwidth]{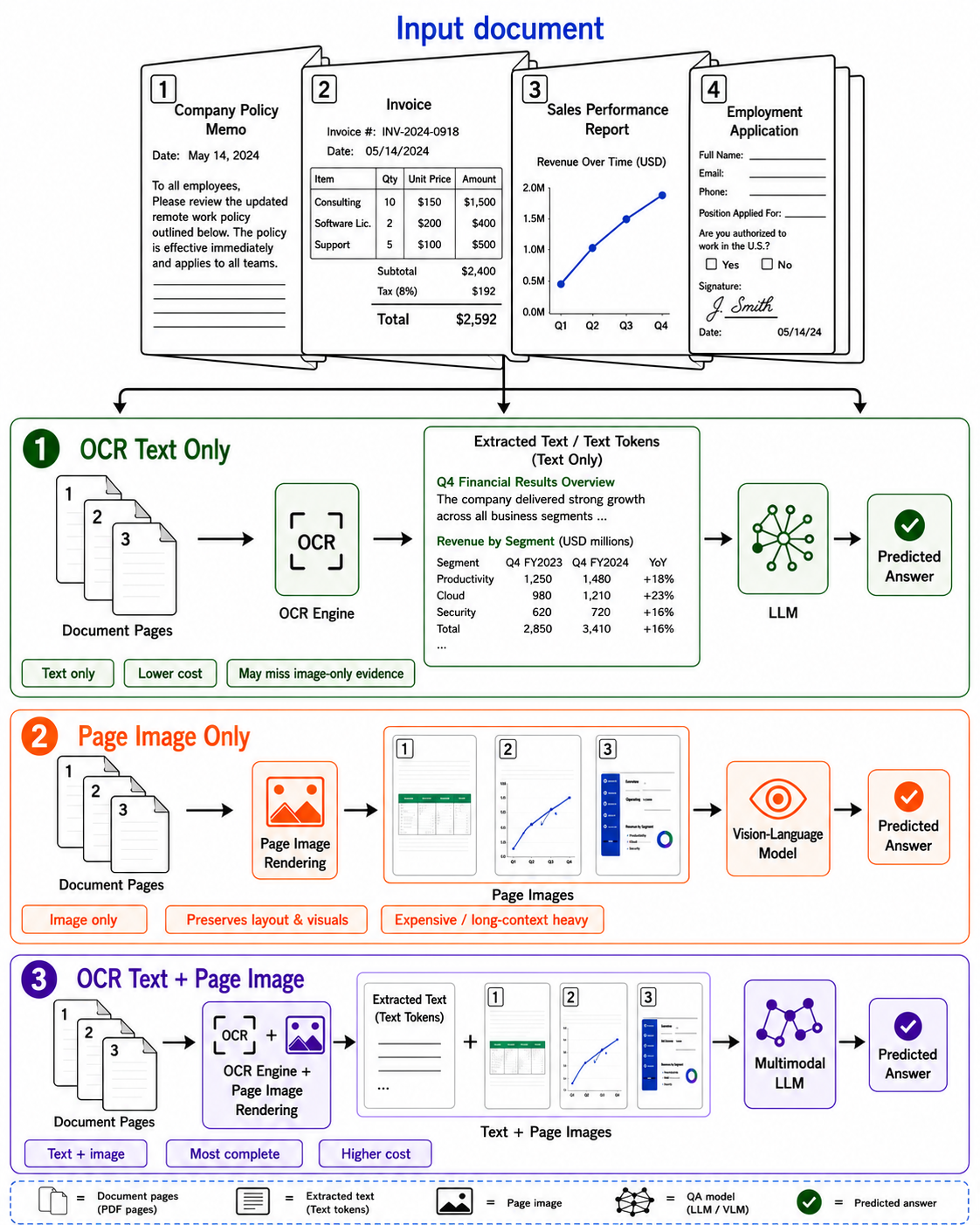}
\caption{Three document representations use the same QA prompt and judge. \vision{} preserves layout and visual content. \textrep{} is compact and uncapped by page count. \both{} concatenates OCR text followed by page images at the cost of larger context.}
\label{fig:overview}
\end{figure}

We address this by comparing \textbf{\vision{}} (page images), \textbf{\textrep{}}
(OCR text), and \textbf{\both{}} (text then images)
across four commercial endpoints from two providers, one public benchmark
\citep{wang2024mmlongbench}, and one enterprise knowledge base. Each model runs
under two retrieval regimes: \textbf{\oracle{}} (gold evidence pages) and
\textbf{\allpages{}} (full document).
Within each model, corpus, and regime, the prompt, judge, and scoring pipeline are held constant; the representation and its resulting input size are the intended experimental differences.

\begin{table*}[t]
\centering
\small
\setlength{\tabcolsep}{5pt}
\caption{Evaluation corpora. The \textbf{$\leq$50-page} columns count the documents short enough to
fit the vision page cap and their questions (the subset used in Table~\ref{tab:results-50p}).}
\label{tab:datasets}
\begin{tabular}{l rr @{\hskip 14pt} rr >{\raggedright\arraybackslash}p{0.46\textwidth}}
\toprule
\rowcolor{headergray}
\textbf{Corpus}
& \textbf{Docs} & \textbf{\shortstack[r]{$\leq$50-page\\Docs}}
& \textbf{Questions} & \textbf{\shortstack[r]{$\leq$50-page\\Questions}}
& \textbf{Profile} \\
\midrule
\textbf{Enterprise KB} & 1{,}447 & 1{,}445 & 5{,}169 & 5{,}163 & Short support articles with screenshots, UI states, diagrams, and cross-modal references; median 4pp., max 237pp.; 99.86\% of docs $\leq$50pp. \\
\addlinespace[2pt]
\textbf{MMLongBench-Doc} & 135 & 101 & 1{,}091 & 798 & Visually rich PDFs: reports, papers, manuals; cross-page and unanswerable questions \citep{wang2024mmlongbench}; median 28pp., max 468pp.\ (mean 48.3); 25\% of docs $>$50pp. \\
\bottomrule
\end{tabular}
\vspace{-0.5em}
\end{table*}

Our contributions:
\begin{itemize}[leftmargin=*,itemsep=1pt,topsep=2pt]
\item A \textbf{controlled representation benchmark} across four models, two corpora, and two retrieval regimes with a fixed prompt and judge.
\item A \textbf{systematic analysis of how document length interacts with input representation},
showing that \vision{} leads on accuracy within the image budget while their latency and
cost grow with document length (Section~\ref{sec:crossover}).
\item A \textbf{lightweight question-text router} that captures much of the accuracy gap to
always-vision while retaining most of text's latency advantage, evaluated on a document-disjoint
held-out split (Section~\ref{sec:router}).
\end{itemize}

\section{Related Work}

\paragraph{Document QA and representation.}
Document QA has expanded from single-page visual understanding \citep{mathew2021docvqa} to multi-page reasoning over tables, charts, and cross-page evidence \citep{van2023document,wang2024mmlongbench,deng2025longdocurl}. Layout-aware models such as LayoutLMv3 \citep{huang2022layoutlmv3} and Pix2Struct \citep{lee2023pix2struct} learn joint text--image representations. \citet{berghaus2025multi} compare image- and text-based processing for invoice extraction and report model- and dataset-dependent results. Our experiments compare representations under a common prompt and evaluation pipeline across four models, two corpora, and two regimes.

\paragraph{Multimodal document retrieval and benchmarks.}
Retrieval-augmented generation \citep{lewis2020retrieval} couples two decisions: which evidence to retrieve and how to represent it. PDFTriage \citep{saad2024pdftriage} routes questions to document structure; M3DocRAG \citep{cho2024m3docrag} and MMDocRAG \citep{dong2026benchmarking} retrieve multimodal document evidence. Our \oracle{}/\allpages{} regimes compare gold-page input with no page retrieval. Long-context position effects are established for text \citep{liu2024lost}, and prompt-compression methods reduce text context length \citep{jiang2023llmlingua,li2023compressing}. We examine how document representation interacts with full-document context and a fixed image budget.
A parallel line of work asks whether page images can replace extracted text directly. ColPali \citep{faysse2024colpali} and Document Screenshot Embedding \citep{ma2024dse} embed rendered pages rather than parsed text, and VisRAG \citep{yu2024visrag} builds a vision-language RAG pipeline over page images, each reporting that image representations avoid parsing loss on layout-rich documents. Our controlled comparison isolates this same text-versus-image choice at generation time rather than at retrieval time.

\begin{table*}[!t]
\centering
\small
\setlength{\tabcolsep}{3pt}
\caption{Evaluation metrics: accuracy as the primary outcome; 
pairwise disagreement, and latency}
\label{tab:metrics}
\begin{tabular}{>{\raggedright\arraybackslash}p{0.20\textwidth}>{\raggedright\arraybackslash}p{0.37\textwidth}>{\raggedright\arraybackslash}p{0.37\textwidth}}
\toprule
\rowcolor{headergray}
\textbf{Metric} & \textbf{Definition} & \textbf{Interpretation} \\
\midrule
\textbf{Accuracy} &
\(\acc(s,m,d,r)=|Q|^{-1}\sum_q c_q\), with \(c_q\in[0,1]\) the mean judged correctness across prediction and judge runs. &
Average judged answer correctness; higher is better. \\
\addlinespace
\addlinespace
\textbf{Pairwise disagreement} &
\(\dis(s_i,s_j)=|Q|^{-1}\sum_q \mathbf{1}[b_{q,i}\neq b_{q,j}]\), where \(b_{q,s}\) is the binary correctness label used for paired analysis. &
Higher values indicate more complementary correctness outcomes. \\
\addlinespace
\textbf{Latency} &
Average seconds per question from run telemetry, by representation and regime. &
High latency can make an accurate route impractical. \\
\bottomrule
\end{tabular}
\end{table*}

\paragraph{Parsing and evaluation.}
Text quality depends on the parsing pipeline. Neural parsers \citep{blecher2024nougat,nassar2025smoldocling} produce structured markup from page images, while direct multimodal processing can retain non-textual evidence that text extraction omits \citep{shen2026ocr}. Our text condition uses a proprietary pipeline that outperforms Tesseract \citep{smith2007overview} on three OCR datasets (Appendix Table~\ref{tab:ocr-comparison}); this comparison does not eliminate extraction error as a possible source of the text--image gap. We use the same LLM judge and prompt in every condition, while recognizing that LLM-based evaluation can itself introduce error \citep{zheng2023judging,liu2023g,bavaresco2025llms}.

\section{Setup}

\paragraph{\normalcolor Representation strategies.}
\vision{} \normalcolor sends each page as a 300-DPI image; \textrep{} sends page text from a proprietary OCR pipeline whose comparison with few open-sourced OCR engines appears in Appendix Table~\ref{tab:ocr-comparison}; \both{} supplies the OCR text followed by the page images. All three use the same QA prompt (Appendix~\ref{app:prompts}).

\paragraph{Datasets.}
Two corpora span short, visually dense articles and longer mixed-format documents (Table~\ref{tab:datasets}). 
The enterprise knowledge base (KB) averages 5.7 pages per document and has 93.6\% of its questions labeled as requiring visual or cross-modal evidence.
MMLongBench-Doc \citep{wang2024mmlongbench} contains longer PDFs (48.3 pages on average) with single-page, cross-page, and unanswerable questions. The KB is internal, whereas MMLongBench-Doc provides a public comparison corpus (Appendix~\ref{app:data}).

\begin{table*}[t]
\centering
\caption{Accuracy on the \textbf{$\leq$50-page subset} (documents short enough that every
representation sees the full page set, i.e.\ no vision-input truncation). Cells show mean $\pm$ std across runs, quantifying endpoint and judge
nondeterminism rather than question-level uncertainty; the \emph{Avg.} row is the arithmetic mean of the four model means
per column. Left block: \oracle{} (gold evidence pages); right block: \allpages{} (full document).}
\label{tab:results-50p}
\footnotesize
\setlength{\tabcolsep}{5pt}
\begin{tabular}{l ccc !{\vrule} ccc}
\toprule
& \multicolumn{3}{c}{\textbf{\oracle{}}} & \multicolumn{3}{c}{\textbf{\allpages{}}} \\
\cmidrule(lr){2-4}\cmidrule(lr){5-7}
\textbf{Model}
& \cellcolor{textbg}\textrep{} & \cellcolor{visionbg}\vision{} & \cellcolor{bothbg}\both{}
& \cellcolor{textbg}\textrep{} & \cellcolor{visionbg}\vision{} & \cellcolor{bothbg}\both{} \\
\midrule
\multicolumn{7}{l}{\textbf{Enterprise KB}}\\
Gemini 3.1 Flash-Lite & $0.576 \pm 0.002$ & \best{0.729 $\pm$ 0.002} & $0.724 \pm 0.001$ & $0.594 \pm 0.001$ & $0.711 \pm 0.003$ & \best{0.714 $\pm$ 0.000} \\
Gemini 3.1 Pro        & $0.618 \pm 0.003$ & \best{0.741 $\pm$ 0.004} & $0.736 \pm 0.002$ & $0.634 \pm 0.003$ & \best{0.728 $\pm$ 0.002} & $0.728 \pm 0.003$ \\
GPT-5.4               & $0.639 \pm 0.003$ & $0.677 \pm 0.002$ & \best{0.733 $\pm$ 0.004} & $0.663 \pm 0.006$ & $0.679 \pm 0.004$ & \best{0.734 $\pm$ 0.005} \\
GPT-5.4-mini          & $0.612 \pm 0.002$ & $0.683 \pm 0.003$ & \best{0.731 $\pm$ 0.002} & $0.628 \pm 0.003$ & $0.672 \pm 0.003$ & \best{0.715 $\pm$ 0.005} \\
\cdashline{1-7}
\emph{Avg.}           & 0.611 & 0.708 & \best{0.731} & 0.630 & 0.698 & \best{0.723} \\
\addlinespace[5pt]
\multicolumn{7}{l}{\textbf{MMLongBench-Doc}}\\
Gemini 3.1 Flash-Lite & $0.529 \pm 0.001$ & \best{0.625 $\pm$ 0.003} & $0.613 \pm 0.002$ & $0.511 \pm 0.001$ & $0.615 \pm 0.009$ & \best{0.620 $\pm$ 0.006} \\
Gemini 3.1 Pro        & $0.641 \pm 0.007$ & \best{0.711 $\pm$ 0.008} & $0.691 \pm 0.003$ & $0.628 \pm 0.007$ & \best{0.690 $\pm$ 0.004} & $0.623 \pm 0.006$ \\
GPT-5.4               & $0.559 \pm 0.007$ & $0.667 \pm 0.004$ & \best{0.673 $\pm$ 0.003} & $0.599 \pm 0.009$ & \best{0.686 $\pm$ 0.008} & $0.683 \pm 0.008$ \\
GPT-5.4-mini          & $0.527 \pm 0.009$ & $0.634 \pm 0.005$ & \best{0.643 $\pm$ 0.007} & $0.547 \pm 0.005$ & $0.633 \pm 0.004$ & \best{0.642 $\pm$ 0.007} \\
\cdashline{1-7}
\emph{Avg.}           & 0.564 & \best{0.659} & 0.655 & 0.571 & \best{0.656} & 0.642 \\
\bottomrule
\end{tabular}
\end{table*}

\paragraph{Models and retrieval regimes.}

We evaluate four models from two providers: Gemini~3.1 Flash-Lite and Gemini~3.1 Pro (Google via Vertex AI), GPT-5.4 and GPT-5.4-mini (OpenAI via Azure)\footnote{Exact API identifiers in Appendix~\ref{app:settings}.}. Each model runs under two regimes that bracket realistic retrieval:
\begin{itemize}[leftmargin=*,itemsep=1pt,topsep=2pt]
\item \textbf{\oracle{}} supplies only the gold evidence pages. It measures answer generation when annotated evidence pages are provided, without evaluating page retrieval.
\item \textbf{\allpages{}} sends the full document with no retrieval. Here the model has to both find the relevant pages and extract the answer from a much larger, noisier context.
\end{itemize}
The gap between regimes measures sensitivity to replacing annotated evidence pages with the full document: a large positive gap indicates that the representation benefits from page selection, whereas a small or negative gap indicates comparable or better performance with full context.

\begin{table*}[t]
\centering
\small
\setlength{\tabcolsep}{4pt}
\caption{Mean latency (seconds/question, $\pm$ std across runs) on the \textbf{$\leq$50-page
subset}, where no representation is truncated by the vision page cap. Left block: \oracle{}
(gold evidence pages); right block: \allpages{} (full document). The \emph{Avg.} row is the
arithmetic mean of the four model means per column. Lower is better.}
\label{tab:latency-50p}
\begin{tabular}{l ccc !{\vrule} ccc}
\toprule
& \multicolumn{3}{c}{\textbf{\oracle{}}} & \multicolumn{3}{c}{\textbf{\allpages{}}} \\
\cmidrule(r){2-4}\cmidrule(l){5-7}
\textbf{Model}
& \cellcolor{textbg}\textrep{} & \cellcolor{visionbg}\vision{} & \cellcolor{bothbg}\both{}
& \cellcolor{textbg}\textrep{} & \cellcolor{visionbg}\vision{} & \cellcolor{bothbg}\both{} \\
\midrule
\multicolumn{7}{l}{\emph{Enterprise KB}}\\
Gemini 3.1 Flash-Lite & $1.29 \pm 0.00$ & $2.42 \pm 0.38$ & $2.52 \pm 0.31$ & $1.40 \pm 0.01$ & $3.26 \pm 0.60$ & $3.20 \pm 0.32$ \\
Gemini 3.1 Pro        & $5.57 \pm 0.25$ & $5.69 \pm 0.36$ & $5.36 \pm 0.37$ & $6.21 \pm 0.25$ & $5.68 \pm 0.15$ & $5.81 \pm 0.22$ \\
GPT-5.4               & $3.46 \pm 0.29$ & $4.97 \pm 0.42$ & $4.48 \pm 0.42$ & $3.42 \pm 0.18$ & $5.43 \pm 1.07$ & $5.71 \pm 0.43$ \\
GPT-5.4-mini          & $1.68 \pm 0.20$ & $2.55 \pm 0.15$ & $2.42 \pm 0.15$ & $1.78 \pm 0.42$ & $3.36 \pm 0.13$ & $3.26 \pm 0.16$ \\
\cdashline{1-7}
\emph{Avg.}           & 3.00 & 3.91 & 3.70 & 3.20 & 4.43 & 4.50 \\
\addlinespace[5pt]
\multicolumn{7}{l}{\emph{MMLongBench-Doc}}\\
Gemini 3.1 Flash-Lite & $1.72 \pm 0.09$ & $3.59 \pm 0.44$ & $4.06 \pm 0.16$ & $2.10 \pm 0.06$ & $5.88 \pm 0.52$ & $6.93 \pm 0.28$ \\
Gemini 3.1 Pro        & $6.10 \pm 0.22$ & $6.88 \pm 0.69$ & $6.80 \pm 0.72$ & $7.57 \pm 0.31$ & $8.40 \pm 0.50$ & $10.12 \pm 2.31$ \\
GPT-5.4               & $4.21 \pm 0.32$ & $8.03 \pm 0.75$ & $7.83 \pm 0.67$ & $4.78 \pm 0.24$ & $13.37 \pm 0.61$ & $13.36 \pm 0.51$ \\
GPT-5.4-mini          & $2.00 \pm 0.23$ & $5.05 \pm 0.36$ & $5.03 \pm 0.31$ & $2.22 \pm 0.27$ & $10.20 \pm 0.31$ & $10.09 \pm 0.36$ \\
\cdashline{1-7}
\emph{Avg.}           & 3.51 & 5.89 & 5.93 & 4.17 & 9.46 & 10.13 \\
\bottomrule
\end{tabular}
\end{table*}

\paragraph{Page cap and document scope.}
Due to model-provider limits, we cap all requests at \textbf{50 page images} and restrict the
main analysis to documents of 50 pages or fewer, so that every representation sees the complete
page set. We adopt this restriction specifically to isolate the page-cap truncation effect from
any genuine modality effect. Applying the same cap to both providers keeps the \vision{}
conditions comparable.
For completeness, Appendix~\ref{app:full-results} reports results over \emph{all} documents,
still capped at 50 page images. In that setting, documents longer than 50 pages are truncated
under \vision{}, while \textrep{} continues to see the full document, and the two effects are
shown to be separable.

\paragraph{Scoring and metrics.}
Models produce free-form answers with a 3{,}000-token limit, temperature~0, and a fixed seed of 30. For
each model, representation, dataset, and regime, we average correctness over \textbf{five
prediction runs}. Each prediction is then evaluated by a fixed GPT-5.2 judge over \textbf{three
judging runs}, and we report the mean score. Per-cell run-to-run standard deviations are reported
alongside every accuracy and latency value in Tables~\ref{tab:results-50p}
and~\ref{tab:latency-50p}. Reported $\pm$ values are standard deviations across the five prediction
and three judging runs. Because decoding is greedy with a fixed seed,
they quantify residual nondeterminism in the hosted endpoints and the
judge rather than sampling uncertainty over questions. Table~\ref{tab:metrics} defines the three metrics.

\begin{figure*}[t]
\centering
\includegraphics[width=0.85\textwidth]{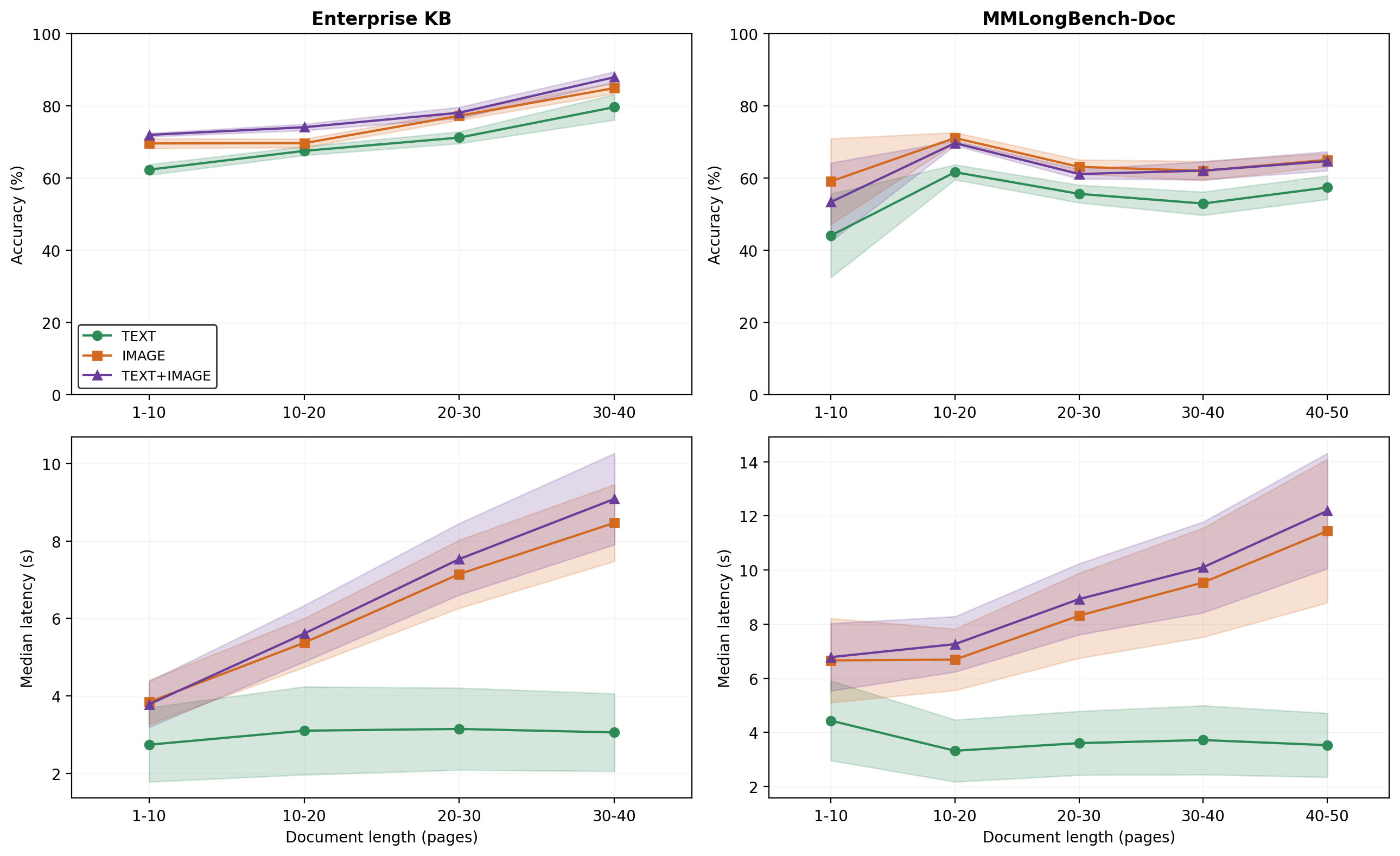}
\caption{Accuracy (top) and median latency (bottom) versus document-length bucket on the
$\leq$50-page subset under \allpages{} retrieval, aggregated across all four models. The Enterprise
KB has no 40--50 page bucket, as its documents are shorter (median 4 pages). Bands show $\pm$ std
across models.}
\label{fig:doc-length-combined}
\end{figure*}

\section{Results}

We organize results around two questions: how the accuracy--latency trade-off changes with
document length, and whether text and images miss the same questions. Table~\ref{tab:results-50p}
and Figure~\ref{fig:doc-length-combined} show that on the $\leq$50-page subset, image-bearing
inputs (\vision{}, \both{}) lead on accuracy across document lengths on both corpora, while
\textrep{} stays consistently lower but at a fraction of the latency: text latency is roughly
flat in document length, whereas \vision{} and \both{} latency grows steadily. The accuracy gap
narrows on the shortest and longest KB buckets but does not close. Section~\ref{sec:router} then
exploits the resulting complementarity with a lightweight router.

\subsection{Accuracy--latency trade-off across representations}
\label{sec:crossover}
Page images retain layout and non-textual content that extracted text may omit
\citep{shen2026ocr}, but any accuracy gain must be weighed against latency. We examine this
trade-off under two regimes, on the $\leq$50-page subset where every representation sees the
complete page set.
\paragraph{Oracle: image inputs lead with gold pages.}
Under \oracle{}, either \vision{} or \both{} is best for every model on both corpora, with gains
over \textrep{} of up to 14.8~percentage points on the KB and 11.6~points on MMLongBench-Doc
(Table~\ref{tab:results-50p}). This is consistent with image inputs retaining evidence omitted by
the text pipeline, but it does not identify whether the difference arises from visual evidence,
OCR errors, or model-specific processing. The gain comes at a latency cost: on the KB, image
input raises average latency from 3.0\,s (\textrep{}) to 3.9\,s (\vision{}) and 3.7\,s (\both{}),
and on MMLongBench-Doc from 3.5\,s to 5.9\,s (Table~\ref{tab:latency-50p}). The larger \oracle{}
latency gap on MMLongBench-Doc is consistent with its questions citing more evidence pages per
query (1.88 vs.\ 1.05 on the KB (Appendix~\ref{app:data})), so each \vision{} or \both{} request carries more page
images.

\paragraph{Full: image accuracy leads, but the latency gap widens.}
Under \allpages{}, image-bearing inputs still lead on accuracy on both corpora: on the KB, \both{}
and \vision{} average 0.72 and 0.70 versus 0.63 for \textrep{}; on MMLongBench-Doc, \vision{}
averages 0.66 versus 0.57 for \textrep{} (Table~\ref{tab:results-50p}). 
Figure~\ref{fig:doc-length-combined} shows \textrep{} latency staying roughly flat in document
length (${\sim}$3\,s on the KB, ${\sim}$3--4\,s on MMLongBench-Doc), while \vision{} and \both{}
latency rise steadily, reaching ${\sim}$9\,s on the longest KB bucket and ${\sim}$12\,s on the
longest MMLongBench-Doc bucket. Accuracy, by contrast, remains image-favored in every length bucket
within the subset. The bucket-level view is included to show that the
accuracy ordering does not reverse within the un-truncated range; the
reversal reported in Appendix~\ref{app:full-results} appears only once the page
cap binds.

\begin{table}[t]
\centering
\small
\setlength{\tabcolsep}{8pt}
\caption{Average cost (USD per 100 questions) and latency (seconds per question) per representation
strategy, averaged across the four models and two datasets. Full per-model, per-dataset
latency breakdowns are in Table~\ref{tab:latency-50p}.}
\label{tab:cost-latency-summary}
\begin{tabular}{l r r}
\toprule
\textbf{Strategy} & \shortstack[r]{Cost\\(\$/100q)} & \shortstack[r]{Latency\\(s/q)} \\
\midrule
\multicolumn{3}{l}{\textbf{\oracle{}}}\\
\textrep{}  & 0.61 & 3.25 \\
\vision{}   & 0.91 & 4.90 \\
\both{}     & 1.29 & 4.81 \\
\addlinespace[2pt]
\midrule
\addlinespace[2pt]

\multicolumn{3}{l}{\textbf{\allpages{}}}\\
\textrep{}  & 1.48 & 3.69 \\
\vision{}   & 2.78 & 6.95 \\
\both{}     & 3.51 & 7.31 \\
\bottomrule
\end{tabular}
\end{table}

\paragraph{Does combining both modalities help?}
\label{sec:both}
Whether \both{} helps depends on the endpoint and corpus. GPT-5.4 and GPT-5.4-mini gain roughly
5~pp (5.6 and 4.8, respectively) from \both{} over \vision{} on the KB under \oracle{}, and their
\vision{}-vs-\both{} disagreement is the highest among the KB cells (13--14\%,
Table~\ref{tab:disagree-50p}), showing that the additional text changes some correctness outcomes.
Gemini~Pro's \vision{} and \both{} outcomes differ less often (${\sim}$7--8\% on the KB). On
MMLongBench-Doc/\allpages{}, Gemini~Pro's \both{} falls behind its \vision{} result and roughly
ties its \textrep{} result (0.623 vs.\ 0.690 and 0.628), while for the GPT endpoints \both{} is the
most accurate condition on the KB. \both{} provides no consistent latency benefit and is the
slowest condition on the longer corpus (Table~\ref{tab:latency-50p}).

\begin{table*}[t]
\centering
\footnotesize
\setlength{\tabcolsep}{3pt}
\caption{Pairwise disagreement rate (\%) on the \textbf{$\leq$50-page subset}: share of questions
where exactly one format is correct (i.e., the sum of text-only-correct and vision-only-correct
questions). A rate of 10\% in the \textrep{}/\vision{} column means that on 10\% of questions one
format answered correctly and the other did not. \colorbox{highdis}{Coral} ${\ge}20\%$;
\colorbox{lowdis}{mint} ${\le}10\%$.}
\label{tab:disagree-50p}
\begin{tabular}{l rr rr @{\hskip 8pt} rr rr}
\toprule
& \multicolumn{4}{c}{\textbf{Enterprise KB}} & \multicolumn{4}{c}{\textbf{MMLongBench-Doc}} \\
\cmidrule(lr){2-5}\cmidrule(lr){6-9}
& \multicolumn{2}{c}{\textrep{}\,/\,\vision{}} & \multicolumn{2}{c}{\vision{}\,/\,\both{}} & \multicolumn{2}{c}{\textrep{}\,/\,\vision{}} & \multicolumn{2}{c}{\vision{}\,/\,\both{}} \\
\cmidrule(lr){2-3}\cmidrule(lr){4-5}\cmidrule(lr){6-7}\cmidrule(lr){8-9}
\textbf{Model} & \oracle{} & \allpages{} & \oracle{} & \allpages{} & \oracle{} & \allpages{} & \oracle{} & \allpages{} \\
\midrule
Gemini 3.1 Flash-Lite & \cellcolor{highdis}24.7 & \cellcolor{highdis}22.2 & 10.7 & 10.4 & \cellcolor{highdis}22.8 & \cellcolor{highdis}23.2 & 12.3 & 13.5 \\
Gemini 3.1 Pro        & \cellcolor{highdis}20.4 & 19.2 & \cellcolor{lowdis}6.7 & \cellcolor{lowdis}8.4 & 19.3 & \cellcolor{highdis}21.2 & 10.4 & 16.0 \\
GPT-5.4               & \cellcolor{highdis}22.9 & 19.7 & 13.5 & 12.7 & 19.7 & \cellcolor{highdis}20.2 & \cellcolor{lowdis}8.9 & \cellcolor{lowdis}7.9 \\
GPT-5.4-mini          & \cellcolor{highdis}24.6 & \cellcolor{highdis}22.6 & 14.0 & 14.2 & \cellcolor{highdis}23.4 & \cellcolor{highdis}20.6 & \cellcolor{lowdis}9.6 & \cellcolor{lowdis}9.5 \\
\bottomrule
\end{tabular}
\end{table*}

\begin{finding}
\textbf{Finding 1.} Document length governs the \emph{cost} of the
trade-off, not its direction. Within the $\leq$50-page subset,
image-bearing inputs (\vision{} or \both{}) lead on
accuracy on both corpora, with no length bucket showing a text
advantage; what grows with length is their latency, which rises
steadily while text latency stays roughly flat.
\end{finding}

\paragraph{The image advantage carries a cost and latency premium.}
Table~\ref{tab:cost-latency-summary} summarizes cost and latency across the four models and two
corpora. Within each regime, \textrep{} is consistently the cheapest and fastest condition:
under \oracle{}, moving from \textrep{} to \vision{} raises cost 1.5$\times$ (0.61 to
0.91~\$/100q) and latency from 3.3 to 4.9~s/q, and \both{} is more expensive still (1.29~\$/100q).

The premium widens under \allpages{}, where the full document is processed: \vision{} and \both{} cost 1.9$\times$ and 2.4$\times$ as much as \textrep{}, respectively, while their latency is roughly twice that of text (7.0 and 7.3~s/q vs.\ 3.7~s/q). Since \vision{} or \both{} also lead on accuracy (Section~\ref{sec:crossover}), no representation dominates on both axes. This trade-off motivates per-question routing, using image input only when it is likely to improve the answer.

\subsection{Text and images answer different questions}

\textrep{} and \vision{} give different correctness outcomes on approximately \textbf{19--25\%}
of questions across the reported corpus, model, and regime cells (Table~\ref{tab:disagree-50p}).
This disagreement decomposes into a text-only-correct slice and a larger vision-only-correct
slice, reflecting vision's higher overall accuracy. On the short KB articles, \oracle{} and
\allpages{} disagreement rates are nearly identical because the full document is already close to
the evidence pages. On MMLongBench-Doc the two regimes diverge somewhat more, consistent with
extra context amplifying representation differences. Because both exclusive-correct slices are
present in every cell, an oracle that picks the correct format per question would beat either
fixed policy.

\begin{finding}
\textbf{Finding 2.} Text and images answer different questions: 19--25\% \textrep{}/\vision{}
disagreement across the reported cells. Neither format subsumes the other; this establishes
headroom for per-question selection.
\end{finding}

\section{Per-Question Representation Routing}
\label{sec:router}

\label{sec:router}
Text and images each answer questions the other misses (Finding~2). This complementarity raises a
concrete question: if a system could decide \emph{per question} whether to send text or images, it
could capture the accuracy of images where they help while keeping text's lower latency and cost elsewhere.
We test whether this is achievable using only the question text, as a proof of concept that
quantifies the potential benefit rather than a production-ready system.

\paragraph{Router design.}

We frame routing as binary classification. For each (question, model) pair under \allpages{}, the
label is \(y=1\) if \vision{} is correct and \textrep{} is incorrect, and \(y=0\) otherwise. When
both representations are correct we default to \textrep{} because it is faster in aggregate; when
both are incorrect we also assign \textrep{}, treating the example as a non-vision-win case. We
exclude \both{} from the router: although often competitive, it adds a second input channel and
closely tracks \vision{} in many settings (Table~\ref{tab:results-50p}), so routing between
\textrep{} and \vision{} targets the primary accuracy--latency trade-off.

Labels are obtained by running both representations on every question. This makes the router a
proof of concept: it measures how much a per-question policy \emph{could} gain if it selected
correctly, not the performance of a deployable classifier. In production, a similar router would be
trained on a smaller labeled sample and applied to new questions. In our training set only 13.2\%
of examples are positive, so \vision{} uniquely beats \textrep{} on a small fraction of questions.

We represent each question with TF-IDF features and train a logistic regression classifier with
balanced class weights, using an approximately 80/20 document-level split stratified by dataset and
model. All questions from the same document stay in one partition, ensuring zero document leakage
and a more reliable estimate of generalization to unseen documents.

All router numbers in Table~\ref{tab:router-results} are computed on this 20\% held-out test partition over the \emph{full corpus} (all documents, image budget still capped at 50 pages) and pooled across the four models, so they are micro-averages over pooled questions rather than the per-model, $\leq$50-page macro-averages in Table~\ref{tab:results-50p}; the two are not directly comparable. Always-vision is correspondingly lower on MMLongBench-Doc here, because the held-out set includes documents beyond the 50-page cap where image inputs are truncated (Appendix~\ref{app:full-results}).

\begin{table}[!t]
\centering
\small
\setlength{\tabcolsep}{5pt}
\caption{Held-out question-text router results under \allpages{}, pooling all four models. The test partition is the 20\% split described in the text.
\emph{Routing accuracy} = accuracy when every question uses the modality
the strategy selects (cf.\ per-model accuracy in Table~\ref{tab:results-50p}).
\emph{Implied med.\ latency} = median wall-clock seconds across the
strategy's per-question routing decisions (cf.\ per-model latency in
Table~\ref{tab:latency-50p}).
Gains are computed from unrounded scores. \textbf{Bold} = best per column within each dataset.}
\label{tab:router-results}
\begin{tabular}{l ccc}
\toprule
\textbf{Strategy} & \textbf{Routing acc.} & \textbf{Gain vs text} & \textbf{Lat.} \\
\midrule
\multicolumn{4}{l}{\textbf{Overall}}\\
Always-text            & 63.6\% & $+$0.0\,pp & \best{2.5\,s} \\
Always-vision          & \best{67.7\%} & $+$4.1\,pp & 4.4\,s \\
TF-IDF router          & 66.2\% & $+$2.6\,pp & 3.1\,s \\
\addlinespace[5pt]
\multicolumn{4}{l}{\textbf{Enterprise KB}}\\
Always-text            & 65.0\% & $+$0.0\,pp & \best{2.4\,s} \\
Always-vision          & \best{69.2\%} & $+$4.2\,pp & 4.1\,s \\
TF-IDF router          & 67.8\% & $+$2.7\,pp & 2.9\,s \\
\addlinespace[5pt]
\multicolumn{4}{l}{\textbf{MMLongBench-Doc}}\\
Always-text            & 53.9\% & $+$0.0\,pp & \best{3.2\,s} \\
Always-vision          & \best{58.0\%} & $+$4.1\,pp & 8.7\,s \\
TF-IDF router          & 55.9\% & $+$2.0\,pp & 5.6\,s \\
\bottomrule
\end{tabular}
\end{table}

\paragraph{Accuracy.}
On the document-disjoint split, always-vision remains the strongest fixed policy, reaching 69.2\% on KB and 58.0\% on MMLongBench-Doc (Table~\ref{tab:router-results}). The router does not beat it, but closes much of the gap over always-text, increasing accuracy by \textbf{2.7\,pp} on KB (65.0 $\to$ 67.8\%) and \textbf{2.0\,pp} on MMLongBench-Doc (53.9 $\to$ 55.9\%) while routing 26.8\% of examples to vision. Overall, the router reaches 66.2\%, gaining \textbf{2.6\,pp} over always-text while remaining 1.5\,pp below always-vision.

\paragraph{Latency.}
Since the router keeps most questions on the faster text path, median latency drops substantially relative to always-vision: from 4.4\,s to 3.1\,s overall, a savings of approximately \textbf{30\%}, while routing only 26.8\% of examples to vision.

\begin{finding}
\textbf{Finding 3.} On the held-out split with no document overlap, a lightweight TF-IDF router gains 2.0–2.7 pp over always-text while cutting median latency by 30\% relative to always-vision. Because training labels require running both representations on every question, this is a proof-of-concept estimate of achievable headroom, not a deployable classifier.
\end{finding}

\section{Discussion}

\paragraph{Document length as the organizing variable.}

Document length organizes the accuracy--latency trade-off, but not by changing which
representation is most accurate. On the $\leq$50-page subset, \vision{} or \both{} leads on
accuracy across all document lengths on both corpora; what scales with length is their cost and
latency, which grow steadily while text stays roughly flat. The apparent accuracy crossover in
favor of text on long documents is a consequence of the fixed 50-image cap truncating image
inputs, and appears only in the full-corpus results (Appendix~\ref{app:full-results}); it is a
property of the current deployment budget, not of the modality. Complementary errors provide a
separate motivation for routing: text and images disagree on roughly a fifth to a quarter of
questions in every cell, so neither representation subsumes the other. The current router,
however, is evaluated only under \allpages{}, and we treat it as a proof of concept rather than a
tuned system.
\paragraph{Multimodal fusion is model-dependent, not additive.}
Sending both text and images does not uniformly help. The GPT endpoints gain approximately 5\,pp
over image-only input on the KB under \oracle{}, whereas the Gemini endpoints show smaller or less
consistent changes (Section~\ref{sec:both}). These outcome differences do not reveal the models'
internal fusion mechanisms, but they show that the preferred representation depends on the endpoint
as well as the document setting.

\paragraph{Implications for RAG pipelines.}
The results motivate evaluating both text and image indices rather than assuming that one representation is uniformly best. The TF-IDF router is a proof of concept; a stronger evaluation would use realistic retrieval, and document features in addition to the question text.

\section{Conclusion}

We compared text, image, and combined document inputs under a common prompting and evaluation pipeline across four model endpoints, two corpora, and two context regimes. Image-bearing inputs outperform text on both corpora, with the largest gains on the short, visually dense Enterprise KB, while text remains cheaper and faster per question. The two representations also produce complementary correctness outcomes, with approximately 19--25\% pairwise disagreement across the reported cells. On a held-out document-level split with no train--test leakage, a lightweight TF-IDF question router recovers 2.0--2.7 percentage points over always-text while reducing median latency by 30\% relative to always-vision.

\section*{Limitations}

First, \oracle{} and \allpages{} do not evaluate a realistic top-$k$ retriever. They compare annotated evidence pages against full-document input, so the results bound two extremes and do not establish performance for intermediate retrieval quality.

Second, we report run-to-run variation but not question-level
uncertainty. Our design fixes the question set and varies only
prediction and judging runs, so the reported standard deviations
capture endpoint and judge nondeterminism. Paired significance testing over the
per-question correctness scores would sharpen these comparisons.

Third, the \textrep{} condition depends on our OCR pipeline. The pipeline outperforms Tesseract on standard document-OCR benchmarks
(Appendix Table~\ref{tab:ocr-comparison}) but is not state of the art, and may extract text less
faithfully than the frontier models themselves would from the same pages. Part of the text--image
gap we report may therefore reflect extraction error rather than an intrinsic limitation of the
text modality.

\section*{Ethics Statement}

The Enterprise KB contains internal product-support documentation owned by the authors' organization and was accessed under existing internal data-use policies. 
Questions were created and reviewed by internal annotators as part of their regular work. All model endpoints, including the judge, were accessed under enterprise agreements with Vertex AI and Azure OpenAI, which exclude submitted prompts and outputs from provider model training. No data was retained beyond the terms of these agreements.
The Enterprise KB and OCR pipeline cannot be released because they are proprietary. We therefore include MMLongBench-Doc to enable independent reproduction of the main text-versus-image comparison.

\bibliography{references}
\appendix

\section{OCR Text Extraction}

We benchmark five OCR systems: \textbf{Tesseract} (v5.5.3)~\cite{smith2007overview}, a widely-used classical baseline; \textbf{PaddleOCR} (v3.7.0)~\cite{cui2025paddleocr30technicalreport} with PP-OCRv5 Latin models; \textbf{EasyOCR} (v1.7.2)~\cite{easyocr} combining CRAFT detection with LSTM/CTC recognition; and \textbf{dots.mocr}~\cite{dotsmocr2026}, a 3B-parameter vision-language model that jointly performs layout detection and text recognition. Our \textbf{in-house engine} is a proprietary CNN-based detector and recognizer trained on public, private, and synthetic data. All systems use GPU inference with their Latin-script configurations.

We evaluate using End-to-End Bag-of-Words (EEBoW) recall, which measures textual completeness independent of spatial accuracy. Text is tokenized into whitespace-delimited words and represented as multisets; recall is the fraction of ground-truth words recovered.

\begin{table}[h]
\centering
\small
\setlength{\tabcolsep}{7pt}
\caption{OCR results on three public benchmarks. End-to-End Bag-of-Words recall.}
\label{tab:ocr-comparison}
\begin{tabular}{lccc}
\toprule
\textbf{Model} & \textbf{CORD-1k} & \textbf{DeepForm} & \textbf{FUNSD} \\
\midrule
Tesseract & 0.474 & 0.668 & 0.589 \\
EasyOCR & 0.511 & 0.855 & 0.428 \\
PaddleOCR & 0.801 & 0.956 & 0.746 \\
dots.mocr & 0.780 & 0.504 & 0.726 \\
In-house & \textbf{0.881} & \textbf{0.970} & \textbf{0.815} \\
\bottomrule
\end{tabular}
\end{table}

Our in-house engine leads across all three public benchmarks (CORD-1k \cite{park2019cord}, DeepForm \cite{svetlichnaya2020deepform}, and FUNSD \cite{jaume2019funsd}), with the largest margin on CORD-1k (0.881 vs.\ 0.474 for Tesseract) shown in Table~\ref{tab:ocr-comparison}. Notably, all benchmarked datasets (Appendix~\ref{app:data}) are inherently multimodal and require visual representations to answer questions. Even with stronger OCR, the gap between text-only and image-based approaches remains large, confirming that visual representations are fundamentally necessary.  Using weaker open-source models would only widen this gap.

\section{Experimental Settings and Provenance}
\label{app:settings}

\paragraph{Input and inference.}

Images use \textbf{300-DPI PDF page renders}. Text uses \textbf{page text}, our proprietary OCR pipeline outperforming other open-source models (Table~\ref{tab:ocr-comparison}). All runs use the same QA prompt (Appendix~\ref{app:prompts}), no learned retriever, seed~30, a \textbf{3{,}000-token answer limit}, and \textbf{at most 50 page images} per request. Temperature is 0 for all four models.

\begin{table*}[t]
\centering
\caption{Effect of restricting to the $\leq$50-page subset. Accuracy is mean~$\pm$~std across
runs; $\Delta = \mathrm{acc}_{\leq 50} - \mathrm{acc}_{\mathrm{all}}$. \best{Bold} $\Delta$
marks $|\Delta|\geq 0.02$. Large positive $\Delta$ concentrates in \vision{}/\allpages{},
reflecting vision-input truncation on documents longer than the page cap.}
\label{tab:results-50p-vs-all}
\scriptsize
\setlength{\tabcolsep}{5pt}
\begin{tabular}{l l l r r r}
\toprule
\textbf{Model} & \textbf{Representation} & \textbf{Regime} & \textbf{Acc$_{\leq 50}$} & \textbf{Acc$_{\mathrm{all}}$} & \textbf{$\Delta$} \\
\midrule
\multicolumn{6}{l}{\textbf{Enterprise KB}}\\
GPT-5.4       & \textrep{} & \allpages{} & $0.6634 \pm 0.0059$ & $0.6626 \pm 0.0059$ & $+0.0008$ \\
GPT-5.4       & \textrep{} & \oracle{}   & $0.6387 \pm 0.0033$ & $0.6382 \pm 0.0033$ & $+0.0005$ \\
GPT-5.4       & \vision{}  & \allpages{} & $0.6792 \pm 0.0043$ & $0.6788 \pm 0.0043$ & $+0.0004$ \\
GPT-5.4       & \vision{}  & \oracle{}   & $0.6774 \pm 0.0018$ & $0.6769 \pm 0.0018$ & $+0.0005$ \\
GPT-5.4       & \both{}    & \allpages{} & $0.7344 \pm 0.0051$ & $0.7338 \pm 0.0052$ & $+0.0006$ \\
GPT-5.4       & \both{}    & \oracle{}   & $0.7329 \pm 0.0039$ & $0.7324 \pm 0.0039$ & $+0.0005$ \\
\addlinespace[3pt]
GPT-5.4-mini  & \textrep{} & \allpages{} & $0.6277 \pm 0.0033$ & $0.6269 \pm 0.0033$ & $+0.0007$ \\
GPT-5.4-mini  & \textrep{} & \oracle{}   & $0.6122 \pm 0.0015$ & $0.6116 \pm 0.0015$ & $+0.0006$ \\
GPT-5.4-mini  & \vision{}  & \allpages{} & $0.6720 \pm 0.0027$ & $0.6714 \pm 0.0027$ & $+0.0006$ \\
GPT-5.4-mini  & \vision{}  & \oracle{}   & $0.6825 \pm 0.0027$ & $0.6821 \pm 0.0028$ & $+0.0005$ \\
GPT-5.4-mini  & \both{}    & \allpages{} & $0.7148 \pm 0.0054$ & $0.7142 \pm 0.0053$ & $+0.0006$ \\
GPT-5.4-mini  & \both{}    & \oracle{}   & $0.7309 \pm 0.0020$ & $0.7304 \pm 0.0020$ & $+0.0005$ \\
\addlinespace[3pt]
Gemini 3.1 Pro & \textrep{} & \allpages{} & $0.6335 \pm 0.0031$ & $0.6330 \pm 0.0031$ & $+0.0005$ \\
Gemini 3.1 Pro & \textrep{} & \oracle{}   & $0.6179 \pm 0.0029$ & $0.6174 \pm 0.0028$ & $+0.0004$ \\
Gemini 3.1 Pro & \vision{}  & \allpages{} & $0.7280 \pm 0.0016$ & $0.7276 \pm 0.0016$ & $+0.0005$ \\
Gemini 3.1 Pro & \vision{}  & \oracle{}   & $0.7409 \pm 0.0035$ & $0.7406 \pm 0.0035$ & $+0.0003$ \\
Gemini 3.1 Pro & \both{}    & \allpages{} & $0.7275 \pm 0.0034$ & $0.7271 \pm 0.0035$ & $+0.0005$ \\
Gemini 3.1 Pro & \both{}    & \oracle{}   & $0.7364 \pm 0.0017$ & $0.7361 \pm 0.0017$ & $+0.0003$ \\
\addlinespace[3pt]
Gemini 3.1 Flash-Lite & \textrep{} & \allpages{} & $0.5942 \pm 0.0009$ & $0.5938 \pm 0.0008$ & $+0.0004$ \\
Gemini 3.1 Flash-Lite & \textrep{} & \oracle{}   & $0.5762 \pm 0.0024$ & $0.5757 \pm 0.0024$ & $+0.0005$ \\
Gemini 3.1 Flash-Lite & \vision{}  & \allpages{} & $0.7112 \pm 0.0025$ & $0.7110 \pm 0.0025$ & $+0.0002$ \\
Gemini 3.1 Flash-Lite & \vision{}  & \oracle{}   & $0.7294 \pm 0.0015$ & $0.7292 \pm 0.0015$ & $+0.0003$ \\
Gemini 3.1 Flash-Lite & \both{}    & \allpages{} & $0.7141 \pm 0.0002$ & $0.7138 \pm 0.0002$ & $+0.0004$ \\
Gemini 3.1 Flash-Lite & \both{}    & \oracle{}   & $0.7243 \pm 0.0010$ & $0.7241 \pm 0.0010$ & $+0.0003$ \\
\midrule
\addlinespace[5pt]
\multicolumn{6}{l}{\textbf{MMLongBench-Doc}}\\
GPT-5.4       & \textrep{} & \allpages{} & $0.5990 \pm 0.0093$ & $0.5927 \pm 0.0086$ & $+0.0063$ \\
GPT-5.4       & \textrep{} & \oracle{}   & $0.5589 \pm 0.0065$ & $0.5615 \pm 0.0077$ & $-0.0026$ \\
GPT-5.4       & \vision{}  & \allpages{} & $0.6857 \pm 0.0078$ & $0.6348 \pm 0.0082$ & \best{$+0.0509$} \\
GPT-5.4       & \vision{}  & \oracle{}   & $0.6672 \pm 0.0043$ & $0.6585 \pm 0.0039$ & $+0.0087$ \\
GPT-5.4       & \both{}    & \allpages{} & $0.6830 \pm 0.0080$ & $0.6636 \pm 0.0079$ & $+0.0193$ \\
GPT-5.4       & \both{}    & \oracle{}   & $0.6734 \pm 0.0027$ & $0.6643 \pm 0.0027$ & $+0.0091$ \\
\addlinespace[3pt]
GPT-5.4-mini  & \textrep{} & \allpages{} & $0.5466 \pm 0.0046$ & $0.5430 \pm 0.0038$ & $+0.0036$ \\
GPT-5.4-mini  & \textrep{} & \oracle{}   & $0.5268 \pm 0.0093$ & $0.5340 \pm 0.0097$ & $-0.0072$ \\
GPT-5.4-mini  & \vision{}  & \allpages{} & $0.6333 \pm 0.0038$ & $0.5923 \pm 0.0025$ & \best{$+0.0410$} \\
GPT-5.4-mini  & \vision{}  & \oracle{}   & $0.6343 \pm 0.0052$ & $0.6279 \pm 0.0025$ & $+0.0065$ \\
GPT-5.4-mini  & \both{}    & \allpages{} & $0.6424 \pm 0.0073$ & $0.6236 \pm 0.0040$ & $+0.0187$ \\
GPT-5.4-mini  & \both{}    & \oracle{}   & $0.6426 \pm 0.0065$ & $0.6354 \pm 0.0054$ & $+0.0072$ \\
\addlinespace[3pt]
Gemini 3.1 Pro & \textrep{} & \allpages{} & $0.6276 \pm 0.0072$ & $0.6053 \pm 0.0084$ & \best{$+0.0223$} \\
Gemini 3.1 Pro & \textrep{} & \oracle{}   & $0.6413 \pm 0.0074$ & $0.6320 \pm 0.0066$ & $+0.0093$ \\
Gemini 3.1 Pro & \vision{}  & \allpages{} & $0.6902 \pm 0.0040$ & $0.6326 \pm 0.0050$ & \best{$+0.0576$} \\
Gemini 3.1 Pro & \vision{}  & \oracle{}   & $0.7113 \pm 0.0076$ & $0.7039 \pm 0.0063$ & $+0.0073$ \\
Gemini 3.1 Pro & \both{}    & \allpages{} & $0.6233 \pm 0.0060$ & $0.6077 \pm 0.0042$ & $+0.0156$ \\
Gemini 3.1 Pro & \both{}    & \oracle{}   & $0.6912 \pm 0.0030$ & $0.6849 \pm 0.0013$ & $+0.0064$ \\
\addlinespace[3pt]
Gemini 3.1 Flash-Lite & \textrep{} & \allpages{} & $0.5109 \pm 0.0006$ & $0.5108 \pm 0.0004$ & $+0.0000$ \\
Gemini 3.1 Flash-Lite & \textrep{} & \oracle{}   & $0.5292 \pm 0.0006$ & $0.5228 \pm 0.0004$ & $+0.0065$ \\
Gemini 3.1 Flash-Lite & \vision{}  & \allpages{} & $0.6145 \pm 0.0093$ & $0.5778 \pm 0.0074$ & \best{$+0.0367$} \\
Gemini 3.1 Flash-Lite & \vision{}  & \oracle{}   & $0.6253 \pm 0.0027$ & $0.6205 \pm 0.0033$ & $+0.0048$ \\
Gemini 3.1 Flash-Lite & \both{}    & \allpages{} & $0.6199 \pm 0.0056$ & $0.6101 \pm 0.0011$ & $+0.0097$ \\
Gemini 3.1 Flash-Lite & \both{}    & \oracle{}   & $0.6128 \pm 0.0018$ & $0.6101 \pm 0.0017$ & $+0.0026$ \\
\bottomrule
\end{tabular}
\end{table*}

\begin{table*}[t]
\centering
\scriptsize
\setlength{\tabcolsep}{5pt}
\caption{Effect of the $\leq$50-page restriction on latency (seconds/question, mean~$\pm$~std);
$\Delta = \mathrm{lat}_{\leq 50} - \mathrm{lat}_{\mathrm{all}}$ (negative = subset is faster).
\best{Bold} $\Delta$ marks $|\Delta|\geq 0.5$s. Large negative $\Delta$ concentrates in
\vision{}/\both{} under \allpages{}: the excluded long documents are the ones that inflate
vision latency.}
\label{tab:latency-50p-vs-all}
\begin{tabular}{l l l r r r}
\toprule
\textbf{Model} & \textbf{Representation} & \textbf{Regime} & \textbf{Lat$_{\leq 50}$} & \textbf{Lat$_{\mathrm{all}}$} & \textbf{$\Delta$ (s)} \\
\midrule
\multicolumn{6}{l}{\textbf{Enterprise KB}}\\
GPT-5.4       & \textrep{} & \allpages{} & $3.42 \pm 0.18$ & $3.42 \pm 0.18$ & $-0.003$ \\
GPT-5.4       & \textrep{} & \oracle{}   & $3.46 \pm 0.29$ & $3.46 \pm 0.29$ & $-0.001$ \\
GPT-5.4       & \vision{}  & \allpages{} & $5.43 \pm 1.07$ & $5.44 \pm 1.08$ & $-0.011$ \\
GPT-5.4       & \vision{}  & \oracle{}   & $4.97 \pm 0.42$ & $4.98 \pm 0.42$ & $-0.009$ \\
GPT-5.4       & \both{}    & \allpages{} & $5.71 \pm 0.43$ & $5.73 \pm 0.43$ & $-0.013$ \\
GPT-5.4       & \both{}    & \oracle{}   & $4.48 \pm 0.42$ & $4.49 \pm 0.42$ & $-0.004$ \\
\addlinespace[3pt]
GPT-5.4-mini  & \textrep{} & \allpages{} & $1.78 \pm 0.42$ & $1.78 \pm 0.42$ & $-0.000$ \\
GPT-5.4-mini  & \textrep{} & \oracle{}   & $1.68 \pm 0.20$ & $1.68 \pm 0.20$ & $-0.001$ \\
GPT-5.4-mini  & \vision{}  & \allpages{} & $3.36 \pm 0.13$ & $3.37 \pm 0.13$ & $-0.011$ \\
GPT-5.4-mini  & \vision{}  & \oracle{}   & $2.55 \pm 0.15$ & $2.55 \pm 0.15$ & $-0.002$ \\
GPT-5.4-mini  & \both{}    & \allpages{} & $3.26 \pm 0.16$ & $3.27 \pm 0.16$ & $-0.012$ \\
GPT-5.4-mini  & \both{}    & \oracle{}   & $2.42 \pm 0.15$ & $2.43 \pm 0.15$ & $-0.000$ \\
\addlinespace[3pt]
Gemini 3.1 Pro & \textrep{} & \allpages{} & $6.21 \pm 0.25$ & $6.21 \pm 0.25$ & $-0.001$ \\
Gemini 3.1 Pro & \textrep{} & \oracle{}   & $5.57 \pm 0.25$ & $5.57 \pm 0.25$ & $-0.001$ \\
Gemini 3.1 Pro & \vision{}  & \allpages{} & $5.68 \pm 0.15$ & $5.68 \pm 0.15$ & $-0.005$ \\
Gemini 3.1 Pro & \vision{}  & \oracle{}   & $5.69 \pm 0.36$ & $5.69 \pm 0.36$ & $+0.000$ \\
Gemini 3.1 Pro & \both{}    & \allpages{} & $5.81 \pm 0.22$ & $5.82 \pm 0.22$ & $-0.009$ \\
Gemini 3.1 Pro & \both{}    & \oracle{}   & $5.36 \pm 0.37$ & $5.36 \pm 0.37$ & $+0.000$ \\
\addlinespace[3pt]
Gemini 3.1 Flash-Lite & \textrep{} & \allpages{} & $1.40 \pm 0.01$ & $1.40 \pm 0.01$ & $-0.001$ \\
Gemini 3.1 Flash-Lite & \textrep{} & \oracle{}   & $1.29 \pm 0.00$ & $1.29 \pm 0.00$ & $-0.000$ \\
Gemini 3.1 Flash-Lite & \vision{}  & \allpages{} & $3.26 \pm 0.60$ & $3.27 \pm 0.60$ & $-0.008$ \\
Gemini 3.1 Flash-Lite & \vision{}  & \oracle{}   & $2.42 \pm 0.38$ & $2.42 \pm 0.38$ & $+0.000$ \\
Gemini 3.1 Flash-Lite & \both{}    & \allpages{} & $3.20 \pm 0.32$ & $3.21 \pm 0.32$ & $-0.007$ \\
Gemini 3.1 Flash-Lite & \both{}    & \oracle{}   & $2.52 \pm 0.31$ & $2.52 \pm 0.31$ & $+0.000$ \\
\midrule
\addlinespace[5pt]
\multicolumn{6}{l}{\textbf{MMLongBench-Doc}}\\
GPT-5.4       & \textrep{} & \allpages{} & $4.78 \pm 0.24$ & $4.99 \pm 0.27$ & $-0.207$ \\
GPT-5.4       & \textrep{} & \oracle{}   & $4.21 \pm 0.32$ & $4.15 \pm 0.30$ & $+0.067$ \\
GPT-5.4       & \vision{}  & \allpages{} & $13.37 \pm 0.61$ & $14.75 \pm 0.65$ & \best{$-1.386$} \\
GPT-5.4       & \vision{}  & \oracle{}   & $8.03 \pm 0.75$ & $7.99 \pm 0.74$ & $+0.044$ \\
GPT-5.4       & \both{}    & \allpages{} & $13.36 \pm 0.51$ & $14.92 \pm 0.52$ & \best{$-1.555$} \\
GPT-5.4       & \both{}    & \oracle{}   & $7.83 \pm 0.67$ & $7.83 \pm 0.65$ & $-0.004$ \\
\addlinespace[3pt]
GPT-5.4-mini  & \textrep{} & \allpages{} & $2.22 \pm 0.27$ & $2.32 \pm 0.30$ & $-0.103$ \\
GPT-5.4-mini  & \textrep{} & \oracle{}   & $2.00 \pm 0.23$ & $1.99 \pm 0.23$ & $+0.013$ \\
GPT-5.4-mini  & \vision{}  & \allpages{} & $10.20 \pm 0.31$ & $11.53 \pm 0.33$ & \best{$-1.332$} \\
GPT-5.4-mini  & \vision{}  & \oracle{}   & $5.05 \pm 0.36$ & $5.07 \pm 0.37$ & $-0.020$ \\
GPT-5.4-mini  & \both{}    & \allpages{} & $10.09 \pm 0.36$ & $11.50 \pm 0.36$ & \best{$-1.405$} \\
GPT-5.4-mini  & \both{}    & \oracle{}   & $5.03 \pm 0.31$ & $5.04 \pm 0.31$ & $-0.014$ \\
\addlinespace[3pt]
Gemini 3.1 Pro & \textrep{} & \allpages{} & $7.57 \pm 0.31$ & $7.82 \pm 0.48$ & $-0.246$ \\
Gemini 3.1 Pro & \textrep{} & \oracle{}   & $6.10 \pm 0.22$ & $6.12 \pm 0.44$ & $-0.015$ \\
Gemini 3.1 Pro & \vision{}  & \allpages{} & $8.40 \pm 0.50$ & $9.45 \pm 1.17$ & \best{$-1.046$} \\
Gemini 3.1 Pro & \vision{}  & \oracle{}   & $6.88 \pm 0.69$ & $7.06 \pm 0.65$ & $-0.180$ \\
Gemini 3.1 Pro & \both{}    & \allpages{} & $10.12 \pm 2.31$ & $11.61 \pm 2.86$ & \best{$-1.492$} \\
Gemini 3.1 Pro & \both{}    & \oracle{}   & $6.80 \pm 0.72$ & $7.01 \pm 0.81$ & $-0.211$ \\
\addlinespace[3pt]
Gemini 3.1 Flash-Lite & \textrep{} & \allpages{} & $2.10 \pm 0.06$ & $2.31 \pm 0.08$ & $-0.209$ \\
Gemini 3.1 Flash-Lite & \textrep{} & \oracle{}   & $1.72 \pm 0.09$ & $1.74 \pm 0.08$ & $-0.022$ \\
Gemini 3.1 Flash-Lite & \vision{}  & \allpages{} & $5.88 \pm 0.52$ & $6.58 \pm 0.57$ & \best{$-0.694$} \\
Gemini 3.1 Flash-Lite & \vision{}  & \oracle{}   & $3.59 \pm 0.44$ & $3.74 \pm 0.41$ & $-0.152$ \\
Gemini 3.1 Flash-Lite & \both{}    & \allpages{} & $6.93 \pm 0.28$ & $8.05 \pm 0.20$ & \best{$-1.119$} \\
Gemini 3.1 Flash-Lite & \both{}    & \oracle{}   & $4.06 \pm 0.16$ & $4.19 \pm 0.13$ & $-0.124$ \\
\bottomrule
\end{tabular}
\end{table*}

\paragraph{Latency measurement.}
Provider-side automatic retries are disabled for all API calls. Reported latency is the wall-clock time of the single call that produced the answer, excluding any internal retry overhead the provider would otherwise add.

\paragraph{Judging.}
\textbf{GPT-5.2} judges answers across \textbf{three runs} (8{,}000-token limit, two parallel calls per batch) \citep{zheng2023judging,liu2023g,bavaresco2025llms}. Questions are batched in fixed document order.

\paragraph{Model Identifiers.} Run identifiers are \texttt{gemini-3.1-flash-lite}, \texttt{gemini-3.1-pro-preview}, \texttt{gpt-5.4}, and \texttt{gpt-5.4-mini}. These are the exact strings passed at inference time.

\section{Dataset Details}
\label{app:data}
\paragraph{MMLongBench-Doc.}
MMLongBench-Doc contains 135 PDFs with 6{,}522 pages, averaging  48.3 pages per document (median 28,
range 9--468) \citep{wang2024mmlongbench}. We use the current 1{,}091-question release. Questions include 485 single-page (44.5\%), 360 cross-page (33.0\%), and 246 unanswerable (22.5\%). On answerable questions, the evidence labels are multi-label: Pure-text 298 (35.3\%), Table 218 (25.8\%), Chart 178 (21.1\%), Image 293 (34.7\%), and Layout 119 (14.1\%). Answer formats include Integer 34.2\%, String 29.5\%, Float 18.9\%, and List 17.3\%. Questions cite 1.88 evidence pages on average (9.08 distinct evidence pages per document).

\paragraph{Enterprise KB (internal).}
The enterprise KB contains 1{,}447 documents (8{,}296 pages; avg.\ 5.7 pages/doc) with 5{,}169 human-reviewed questions (avg.\ 3.6 questions/doc). Answer mode: visual 3{,}314 (64.1\%), cross-modal 1{,}527 (29.5\%), textual 328 (6.3\%). Difficulty: easy 3{,}731 (72.2\%), medium 1{,}425 (27.6\%), hard 13 (0.3\%). 93.6\% of questions require visual or cross-modal interpretation. Questions cite 1.05 evidence pages on average (2.42 distinct evidence pages per document).

\section{Full-corpus results}
\label{app:full-results}

\begin{figure*}[t]
\centering
\includegraphics[width=0.85\textwidth]{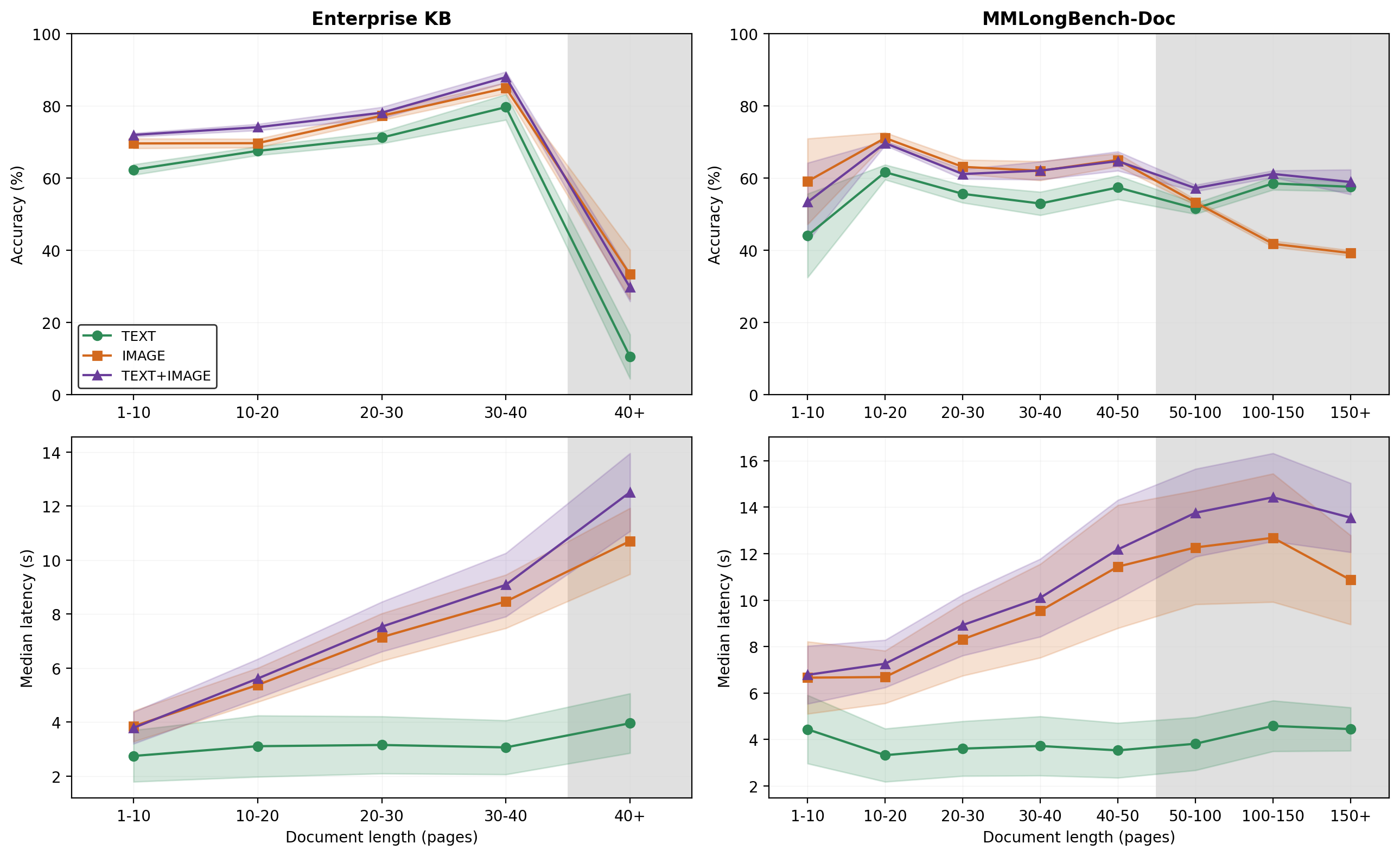}
\caption{Full-corpus accuracy (top) and median latency (bottom) versus document-length bucket
under \allpages{} retrieval, aggregated across all four models. The shaded region marks buckets
above the 50-page image cap, where \vision{} and \both{} inputs are truncated while \textrep{}
sees the full document. Within the un-truncated range, image-bearing inputs lead on accuracy at
every length; the crossover in favor of \textrep{} appears only in the truncated buckets, and
image latency falls there as truncation caps the number of page images processed. Bands show
$\pm$ std across models.}
\label{fig:doc-length-appendix}
\end{figure*}

This appendix reports the per-model results underlying the main text over \emph{all} documents,
with the image budget still capped at 50 pages. Documents longer than 50 pages are therefore
truncated under \vision{} and \both{}, while \textrep{} sees the full document. We report the
effect of the $\leq$50-page restriction on accuracy (Table~\ref{tab:results-50p-vs-all}) and
latency (Table~\ref{tab:latency-50p-vs-all}), the full length-bucket breakdown extending past the
cap (Figure~\ref{fig:doc-length-appendix}).

\paragraph{The restriction isolates the page-cap truncation effect.}

On the Enterprise KB, restricting to $\leq$50 pages changes every accuracy and latency cell by
less than $0.001$ (Tables~\ref{tab:results-50p-vs-all}, \ref{tab:latency-50p-vs-all}): only 6 of
5{,}169 questions come from longer documents, so the subset and the full corpus are effectively
the same measurement. On MMLongBench-Doc, where 25\% of documents exceed the cap, the restriction
has a clear and localized effect. The accuracy gain from restricting concentrates entirely in
\vision{}/\allpages{}, where it reaches $+0.041$ to $+0.058$ across the four models, and is an
order of magnitude smaller for \textrep{}/\allpages{} and for every \oracle{} cell. The latency
change mirrors this: the large reductions ($-1.0$ to $-1.6$\,s) fall on \vision{} and \both{}
under \allpages{}, while \textrep{} and all \oracle{} cells barely move. Both patterns identify the
same cause: only the image conditions on long documents are affected by the restriction, because
only they are truncated by the page cap.

\begin{table}[ht!]
\centering
\scriptsize
\setlength{\tabcolsep}{1pt}
\caption{Non-exclusive error-category shares (\% of incorrect answers) by model, representation,
and regime on Enterprise KB. Left block: \oracle{} (gold evidence pages); right block:
\allpages{} (full document). Columns need not sum to 100\%.}
\label{tab:errors-kb}
\begin{tabular}{l ccc !{\vrule} ccc}
\toprule
& \multicolumn{3}{c}{\textbf{\oracle{}}} & \multicolumn{3}{c}{\textbf{\allpages{}}} \\
\cmidrule(r){2-4}\cmidrule(l){5-7}
\textbf{Model}
& \cellcolor{textbg}\textrep{} & \cellcolor{visionbg}\vision{} & \cellcolor{bothbg}\both{}
& \cellcolor{textbg}\textrep{} & \cellcolor{visionbg}\vision{} & \cellcolor{bothbg}\both{} \\
\midrule
\multicolumn{7}{l}{\emph{partial\_correctness}} \\
Gemini 3.1 Flash-Lite  & 44.9 & 53.0 & 52.9 & 46.0 & 52.0 & 52.1 \\
Gemini 3.1 Pro         & 50.5 & 56.8 & 56.0 & 49.5 & 52.0 & 52.0 \\
GPT-5.4                & 51.5 & 49.1 & 54.7 & 52.4 & 45.5 & 52.0 \\
GPT-5.4-mini           & 49.4 & 48.1 & 53.7 & 51.0 & 49.3 & 51.5 \\
\midrule
\addlinespace
\multicolumn{7}{l}{\emph{extra\_incorrect\_information}} \\
Gemini 3.1 Flash-Lite  & 18.7 & 18.6 & 19.8 & 21.2 & 23.3 & 25.5 \\
Gemini 3.1 Pro         & 18.1 & 17.7 & 20.4 & 20.4 & 22.9 & 22.4 \\
GPT-5.4                & 20.8 & 19.9 & 22.4 & 22.6 & 21.1 & 23.7 \\
GPT-5.4-mini           & 21.8 & 18.5 & 21.6 & 23.8 & 20.8 & 24.4 \\
\midrule
\addlinespace
\multicolumn{7}{l}{\emph{contradiction}} \\
Gemini 3.1 Flash-Lite  & 19.1 & 18.3 & 17.1 & 19.2 & 19.6 & 21.4 \\
Gemini 3.1 Pro         & 16.3 & 16.9 & 15.8 & 19.9 & 19.2 & 19.9 \\
GPT-5.4                & 21.9 & 23.1 & 19.0 & 20.6 & 23.0 & 20.2 \\
GPT-5.4-mini           & 21.7 & 21.3 & 19.5 & 21.9 & 21.1 & 20.0 \\
\midrule
\addlinespace
\multicolumn{7}{l}{\emph{answered\_when\_unanswerable}} \\
Gemini 3.1 Flash-Lite  & 20.7 & 10.4 & 10.8 & 17.7 & 6.6 & 4.9 \\
Gemini 3.1 Pro         & 14.8 & 8.3 & 9.8 & 15.4 & 10.6 & 9.3 \\
GPT-5.4                & 9.3 & 8.3 & 6.1 & 9.6 & 9.1 & 7.2 \\
GPT-5.4-mini           & 9.6 & 7.3 & 4.7 & 9.3 & 6.6 & 4.9 \\
\midrule
\addlinespace
\multicolumn{7}{l}{\emph{numerical\_imprecision}} \\
Gemini 3.1 Flash-Lite  & 6.0 & 4.6 & 5.3 & 5.9 & 5.6 & 5.9 \\
Gemini 3.1 Pro         & 4.8 & 4.0 & 4.1 & 4.4 & 4.0 & 4.2 \\
GPT-5.4                & 6.9 & 10.3 & 6.8 & 6.2 & 10.3 & 5.7 \\
GPT-5.4-mini           & 8.4 & 11.7 & 7.9 & 7.1 & 10.6 & 7.5 \\
\bottomrule
\end{tabular}
\end{table}

\begin{table}[ht!]
\centering
\scriptsize
\setlength{\tabcolsep}{1pt}
\caption{Non-exclusive error-category shares (\% of incorrect answers) by model, representation,
and regime on MMLongBench-Doc. Left block: \oracle{} (gold evidence pages); right block:
\allpages{} (full document). Columns need not sum to 100\%.}
\label{tab:errors-mmlongbench}
\begin{tabular}{l ccc !{\vrule} ccc}
\toprule
& \multicolumn{3}{c}{\textbf{\oracle{}}} & \multicolumn{3}{c}{\textbf{\allpages{}}} \\
\cmidrule(r){2-4}\cmidrule(l){5-7}
\textbf{Model}
& \cellcolor{textbg}\textrep{} & \cellcolor{visionbg}\vision{} & \cellcolor{bothbg}\both{}
& \cellcolor{textbg}\textrep{} & \cellcolor{visionbg}\vision{} & \cellcolor{bothbg}\both{} \\
\midrule
\multicolumn{7}{l}{\emph{partial\_correctness}} \\
Gemini 3.1 Flash-Lite  & 9.6 & 11.7 & 11.7 & 10.1 & 10.8 & 10.6 \\
Gemini 3.1 Pro         & 9.3 & 11.7 & 10.5 & 10.8 & 8.5 & 10.6 \\
GPT-5.4                & 12.3 & 10.4 & 10.2 & 8.7 & 7.6 & 9.1 \\
GPT-5.4-mini           & 9.0 & 9.2 & 8.8 & 11.0 & 8.8 & 9.3 \\
\midrule
\addlinespace
\multicolumn{7}{l}{\emph{extra\_incorrect\_information}} \\
Gemini 3.1 Flash-Lite  & 4.0 & 5.5 & 6.5 & 6.0 & 5.7 & 8.2 \\
Gemini 3.1 Pro         & 4.4 & 5.7 & 4.4 & 8.2 & 6.7 & 7.8 \\
GPT-5.4                & 6.1 & 5.3 & 5.5 & 5.8 & 5.1 & 5.5 \\
GPT-5.4-mini           & 5.8 & 4.0 & 5.9 & 5.4 & 7.0 & 6.6 \\
\midrule
\addlinespace
\multicolumn{7}{l}{\emph{contradiction}} \\
Gemini 3.1 Flash-Lite  & 6.7 & 8.6 & 8.4 & 14.0 & 8.9 & 8.9 \\
Gemini 3.1 Pro         & 8.8 & 9.2 & 7.0 & 8.7 & 7.5 & 7.1 \\
GPT-5.4                & 9.0 & 8.6 & 9.1 & 11.4 & 9.8 & 10.5 \\
GPT-5.4-mini           & 10.8 & 8.2 & 9.5 & 12.2 & 11.1 & 9.5 \\
\midrule
\addlinespace
\multicolumn{7}{l}{\emph{answered\_when\_unanswerable}} \\
Gemini 3.1 Flash-Lite  & 51.6 & 40.4 & 39.0 & 36.3 & 38.2 & 36.5 \\
Gemini 3.1 Pro         & 52.9 & 42.9 & 45.3 & 52.7 & 44.3 & 42.4 \\
GPT-5.4                & 43.6 & 46.8 & 44.6 & 40.5 & 47.2 & 39.7 \\
GPT-5.4-mini           & 44.0 & 45.5 & 42.8 & 37.1 & 41.4 & 37.7 \\
\midrule
\addlinespace
\multicolumn{7}{l}{\emph{numerical\_imprecision}} \\
Gemini 3.1 Flash-Lite  & 30.5 & 35.6 & 37.4 & 37.6 & 39.3 & 40.7 \\
Gemini 3.1 Pro         & 26.5 & 32.1 & 33.6 & 24.9 & 35.8 & 34.8 \\
GPT-5.4                & 32.4 & 31.6 & 33.1 & 36.7 & 32.1 & 36.1 \\
GPT-5.4-mini           & 33.7 & 33.9 & 34.7 & 38.6 & 35.3 & 40.6 \\
\bottomrule
\end{tabular}
\end{table}

\paragraph{The text advantage appears only past the cap.}
Figure~\ref{fig:doc-length-appendix} extends the length-bucket view to the full corpus (shaded
region marks buckets above the 50-page cap). Within the un-truncated range, image-bearing inputs
lead at every length, matching the main text. The crossover in favor of \textrep{} emerges only in
the truncated buckets: on MMLongBench-Doc, \vision{} accuracy falls to ${\sim}$42\% at 100--150
pages and ${\sim}$39\% beyond 150, while \textrep{} holds near 58\%. This is the behavior that
motivates reporting the $\leq$50-page subset in the main text: the crossover is a property of the
fixed image budget, not of the modalities, and it disappears once every representation sees the
complete page set.

\section{Error Category Analysis}

Tables~\ref{tab:errors-kb} and \ref{tab:errors-mmlongbench} break down errors by model, representation, and regime (tagged by judge; Appendix~\ref{app:prompts}). Categories are non-exclusive.

On the KB, \texttt{partial\_correctness} is the most frequent reported category (44--57\%),
\texttt{extra\_incorrect\_information} follows (18--25\%), and switching from
\oracle{} to \allpages{} barely changes the profile. On MMLongBench-Doc,
\texttt{answered\_when\_unanswerable} is high (36--53\%), consistent with the fact that 22.5\% of
questions are unanswerable by design; this category means the model \emph{gave}
an answer to an unanswerable question (over-answering), and its rate is largely
stable across regimes rather than growing with the \oracle{}$\to$\allpages{}
shift. \texttt{numerical\_imprecision} is also frequent (25--41\%); Integer and Float are common answer formats in this benchmark.

\onecolumn

\section{Prompt Templates}
\label{app:prompts}

\subsection{QA Prompt}

All conditions use the prompt below; \vision{} and \both{} additionally append the rendered page images after the text context block. 

\begin{promptbox}[QA Prompt Template]
\begin{lstlisting}[style=prompt]
You are an expert question answering
assistant. Your task is to provide a concise
and accurate answer to the user's question,
strictly based on the provided context.

Instructions:
1. Read the question and the provided context
   carefully.
2. In the "explanation" field, think
   step-by-step to determine how the context
   answers the question.
3. Formulate an answer that directly addresses
   the question.
4. Your answer must be derived solely from the
   information present in the context. Do not
   use outside knowledge or invent facts.
5. If the context does not contain enough
   information, state "Not Answerable".
6. Keep the answer as short as possible:
   a single value, phrase, or list of items.

Output Format:
{
  "explanation": "<reasoning>",
  "answer": "<concise answer>"
}

Question: {{question}}
Context:
\end{lstlisting}
\end{promptbox}

\subsection{LLM-as-Judge Prompt}

The judge receives batches of question-answer pairs through the \texttt{\{\{qa\_pairs\}\}} placeholder, following reference-guided evaluation protocols for open-ended model outputs \citep{zheng2023judging,liu2023g,bavaresco2025llms}.

\begin{promptbox}[LLM-as-Judge Prompt Template]
\begin{lstlisting}[style=prompt]
You are an expert evaluator assessing answers
in a Question and Answer system. For each
question-answer pair you receive:
1. A unique question ID
2. The question text
3. The predicted answer
4. The ground truth answer
Determine:
1. Whether the predicted answer is
   semantically equivalent to the ground truth.
2. All applicable error or warning categories.
Evaluation Instructions:
- Evaluate based on semantic meaning, not
  exact wording.
- Be flexible with phrasing, terminology,
  units, abbreviations, and formatting.
- Mark incorrect only if: (1) core meaning
  differs, (2) contains factually incorrect
  information, (3) contradicts the ground
  truth, or (4) omits key information.
- For numeric values: accept more-precise
  expansions of rounded ground truths; penalize
  rounding that loses accuracy.
- For lists: every item must match; order does
  not matter.
- For multiple-choice: accept the option
  letter, text, or both.
ERROR CATEGORIES (incorrect answers only):
  partial_correctness,
  extra_incorrect_information, contradiction,
  overly_vague, incorrect_yes_no,
  answered_when_unanswerable,
  numerical_imprecision, minor_spelling_error
WARNING CATEGORIES (correct answers only):
  ground_truth_rounding,
  ground_truth_spelling_error,
  redundant_information, extra_information,
  missing_information, other
Output Format:
{ "evaluations": [
    { "question_id": "<id>",
      "is_correct": true/false,
      "explanation": "<pred> vs <truth>",
      "errors": [{"class": "...",
                   "explanation": "..."}],
      "warnings": [{"class": "...",
                     "explanation": "..."}]
    }, ...
] }
\end{lstlisting}
\end{promptbox}

\end{document}